\documentclass[sigconf]{acmart}

\AtBeginDocument{%
  }

\usepackage{multirow}
\usepackage{bm}
\usepackage{algorithm}
\usepackage{algorithmic}
\usepackage{subcaption}
\usepackage{colortbl}
\usepackage{enumitem}

\copyrightyear{2026}
\acmYear{2026}
\setcopyright{cc}
\setcctype{by}
\acmConference[KDD '26]{Proceedings of the 32nd ACM SIGKDD Conference on Knowledge Discovery and Data Mining V.2}{August 09--13, 2026}{Jeju Island, Republic of Korea}
\acmBooktitle{Proceedings of the 32nd ACM SIGKDD Conference on Knowledge Discovery and Data Mining V.2 (KDD '26), August 09--13, 2026, Jeju Island, Republic of Korea}
\acmDOI{10.1145/3770855.3817991}
\acmISBN{979-8-4007-2259-2/2026/08}

\begin{document}

\title{StepFinder: A Temporal Semantic Framework for Failure Attribution in Multi-Agent Systems}

\author{Taiyu Zhu}
\affiliation{%
  \institution{Peking University}
  \city{Beijing}
  \country{China}
}
\email{202221020155@stu.zuel.edu.cn}

\author{Yifan Wu}
\authornote{Corresponding author.}
\affiliation{%
  \institution{Peking University}
  \city{Beijing}
  \country{China}
}
\email{yifanwu@pku.edu.cn}

\author{Weilin Jin}
\affiliation{%
  \institution{Peking University}
  \city{Beijing}
  \country{China}
}
\email{2401112012@stu.pku.edu.cn}

\author{Ying Li}
\authornotemark[1]
\affiliation{%
  \institution{Peking University}
  \city{Beijing}
  \country{China}
}
\email{li.ying@pku.edu.cn}

\author{Gang Huang}
\affiliation{%
  \institution{Peking University}
  \city{Beijing}
  \country{China}
}
\email{hg@pku.edu.cn}

\renewcommand{\shortauthors}{Taiyu Zhu, Yifan Wu, Weilin Jin, Ying Li, and Gang Huang}

\begin{abstract}
  LLM-based multi-agent systems exhibit remarkable collaborative capabilities in complex multi-step tasks. However, these systems are highly sensitive to single-step execution errors that can propagate through agent interactions and lead to cascading failures. To understand the causes of failure and improve system reliability, failure attribution has been introduced as a task that aims to automatically identify the root cause step responsible for a failure. Existing failure attribution methods mainly rely on LLMs to reason over original execution trajectories, which not only incur high inference costs and latency, but also suffer from interference caused by redundant and noisy execution logs, causing LLMs to struggle in accurately identifying the true root cause step.
  To address this, we propose \textbf{StepFinder}, a lightweight failure attribution framework. We use LLMs solely during the feature construction phase to encode execution logs into temporal semantic sequences. Subsequently, a parameter-efficient combination of temporal modeling and attention modules is applied to capture the sequential evolution and cross-step dependencies of the trajectories. Finally, the step-level error score is refined through multi-scale differences and position bias, enabling precise root cause identification.
  Experimental results on the Who\&When benchmark demonstrate that StepFinder outperforms LLM-based methods in step-level failure attribution while achieving substantially higher inference efficiency, reducing inference time by 79\% compared with the fastest LLM-based method, with no text generation overhead. Our code is available at \url{https://github.com/taiyu-zhu/StepFinder}.
\end{abstract}

\ccsdesc[500]{Computing methodologies~Multi-agent systems}
\ccsdesc[500]{Computing methodologies~Neural networks}
\ccsdesc[300]{Computing methodologies~Multi-task learning}
\ccsdesc[500]{Information systems~Information systems applications}

\keywords{Multi-Agent Systems; Failure Attribution; Execution Trace Mining; Temporal Semantic Modeling}

\maketitle

\newcommand\kddavailabilityurl{https://doi.org/10.5281/zenodo.20432323}
\ifdefempty{\kddavailabilityurl}{}{
\begingroup\small\noindent\raggedright\textbf{Resource Availability:}\\
The source code and data of this paper has been made publicly available at \url{\kddavailabilityurl}.
\endgroup
}

\section{Introduction}

\begin{figure}[t]
    \centering
    \includegraphics[width=\columnwidth]{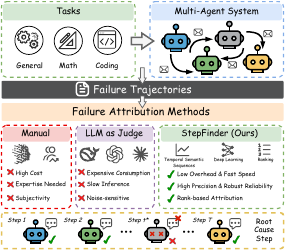} 
    \caption{Overview of MAS failure attribution and comparison of methodologies. This figure illustrates the diagnostic workflow in MAS and provides a multi-dimensional comparison between existing approaches and our proposed model.}
    \label{fig:introduction}
    \Description{Introduction.}
\end{figure}

With the rapid advancement of large language models (LLMs) in text understanding, reasoning, and multi-step planning, they have evolved into autonomous agents capable of decision-making and action execution in complex interactive environments~\cite{brown2020language, wei2022chain, lu2024ai, yao2023tree}. While single LLM-based agents can independently perform tasks such as document analysis, question answering, and tool invocation~\cite{wang2023voyager, erdogan2025plan, yang2024embodied, schick2023toolformer}, they struggle with heterogeneous information integration, long-horizon reasoning, and multi-objective coordination~\cite{valmeekam2023planning, liu2024lost}. To address these limitations, multi-agent systems (MAS) have emerged as a superior paradigm, enabling task decomposition through role specialization and adaptive communication architectures~\cite{wu2024autogen, du2024expressive, zhang2025multi, zhang2023ecoassistant, hou2025model, liu2023dynamic}. By decoupling planning, reasoning, and execution across specialized agents, MAS achieve greater stability and scalability than single-agent systems, especially in long-horizon collaborative tasks~\cite{qiu2024towards, zhang2025towards}. Consequently, MAS consistently outperform single-agent systems in complex scenarios such as program synthesis and scientific discovery~\cite{ghafarollahi2025sciagents, schmidgall2025agent, li2023camel, qian2024chatdev}.

Despite their strong performance on complex tasks, MAS exhibit substantial internal fragility. Empirical studies report failure rates as high as 41\%–86.7\% across representative benchmarks, largely stemming from limitations in system design, task decomposition, and inter-agent coordination~\cite{cemri2025multi}. Moreover, MAS are highly sensitive to minor agent-level deviations, which can be amplified through inter-agent interactions and propagate into cascading system-level failures or severe task drift~\cite{zhang2025breaking}.

Given the high failure risk in MAS, accurate failure attribution is essential for system stability. Traditional efforts have primarily focused on developing fine-grained benchmarks to facilitate manual diagnostic processes~\cite{zhuge2024agent, jimenez2023swe}, yet root cause localization remains a labor-intensive task requiring extensive domain expertise. While recent studies have explored LLM-based automated diagnosis and achieved promising agent-level results~\cite{zhang2025agent, zhang2025agentracer, zhu2025raffles, west2025abduct, banerjee2025did, kong2025aegis}, these approaches typically incur prohibitive computational costs and lack sufficient step-level localization accuracy. As shown in Figure~\ref{fig:introduction}, existing methodologies face a fundamental trade-off between diagnostic depth and operational efficiency, limiting their effectiveness in complex environments.

To address these limitations, we propose \textbf{StepFinder}, an efficient framework for step-level failure attribution in MAS. Our approach frames failure attribution as a structured process across three integrated stages. First, we transform execution trajectories into temporal semantic sequences by encoding the agent identity and action content of each step into dense embeddings. Subsequently, we employ a hybrid architecture combining temporal modeling with agent-aware interactions to capture long-range dependencies and causal links between agents. Each step is assigned an error score refined by multi-scale differences and position bias to pinpoint the root cause. Finally, the model is optimized using a supervised classification loss supplemented by a self-supervised auxiliary loss for future-step prediction, which collectively strengthens trajectory representation and training stability.

To evaluate StepFinder, we conduct extensive experiments on the Who\&When benchmark~\cite{zhang2025agent}, comparing it with three categories of methods: random, LLM-based, and sequential models. Results show that StepFinder consistently outperforms all baselines in attribution precision and ranking quality. Specifically, our model achieves a significant performance leap, improving accuracy by 4.76\% on the Algorithm-Generated subset and 10.35\% on the Hand-Crafted subset compared to the strongest baselines. StepFinder also exhibits superior efficiency, achieving a nearly 5x speedup on the Algorithm-Generated subset by reducing the inference time of the fastest baseline from 2.92s to 0.61s per trajectory.

To summarize, we make the following contributions:
\begin{itemize}[wide=\parindent, leftmargin=*]
    \item We reformulate MAS failure attribution as a structured temporal modeling problem. This perspective allows us to shift the heavy reasoning burden from LLMs to a specialized deep learning architecture.
    \item We propose StepFinder, a lightweight temporal semantic framework designed to identify the root cause step by assigning an explicit error score to each step.
    \item Experiments demonstrate that StepFinder achieves superior attribution precision and ranking quality while significantly reducing inference latency and operational costs compared to existing baselines.
\end{itemize}

\section{Related Work}

\subsection{LLM-based MAS}

In complex, multi-step tasks, single-agent systems can perform well in certain scenarios~\cite{wei2022chain, wang2022self, yao2022react, zheng2023progressive}. However, they exhibit inherent limitations when tasks require collaboration and division of labor. To address this, recent studies have proposed LLM-based MAS, in which agents are assigned distinct roles, structured communication mechanisms are established, and capabilities such as tool usage, memory management, and planning are integrated to enable efficient task decomposition and coordination~\cite{qian2024chatdev, hong2023metagpt, park2023generative}. Early MAS largely relied on manually designed roles, prompts, and communication topologies~\cite{li2023camel, wu2024autogen, zhou2025multi}, limiting system scalability and adaptability. More recent work explores automated role assignment and dynamic interaction topologies, allowing agents to collaborate adaptively during execution, thereby improving task completion efficiency and generality~\cite{li2025assemble, leong2025amas, yang2025agentnet}. These developments indicate that LLM-driven MAS effectively overcome the shortcomings of single-agent systems through multi-agent division of labor and collaboration, providing a scalable solution for intelligent decision-making and multi-step reasoning in complex scenarios.

\subsection{Fragility and Error Patterns in MAS}

Although LLM-based MAS demonstrate strong capabilities, their high error rates and system fragility still limit practical applications~\cite{cemri2025multi, huang2024resilience, owotogbe2025assessing}. During multi-agent collaboration, a single step or individual error can quickly propagate through communication, tool usage, and collaborative dependencies, leading to cascading failures that threaten the reliability of downstream tasks~\cite{zhang2025breaking, gao2025single, shen2025understanding}. Studies indicate that MAS errors are diverse and structured, encompassing cognitive and reasoning mistakes, interaction and collaboration conflicts, planning and execution deviations, as well as system and validation failures. These errors often interact with one another, increasing the complexity of debugging and safeguarding the system~\cite{cemri2025multi, deshpande2025trail, zhu2025llm}. Therefore, a thorough understanding of MAS error mechanisms and patterns, along with incorporating robustness constraints in system design, is crucial for improving the reliability of MAS.

\subsection{Failure Attribution for MAS}

Traditional MAS failure analysis primarily relies on manual or semi-automated annotation, often focused on developing fine-grained metrics for human reference \cite{zhuge2024agent, jimenez2023swe}. However, these processes are inherently time-consuming, subjective, and difficult to scale to long execution traces. Consequently, recent studies have shifted toward automated failure attribution to identify critical error steps and enable systematic diagnosis.

The Who\&When benchmark~\cite{zhang2025agent} holds a pioneering position in this area, formally introducing the task of automated failure attribution for LLM-based MAS. By leveraging the concept of “Decisive Error”, it formalizes the attribution objective as the automatic identification of the responsible agent and the earliest error step, while also providing a standardized benchmark of failure traces to establish a unified evaluation foundation for subsequent research. Building on this, existing studies in automated diagnosis can be broadly categorized into two paradigms. 
The first paradigm, which currently represents the dominant research trend, comprises LLM-based diagnostic methods that directly leverage the model’s reasoning capabilities for failure attribution~\cite{zhang2025agent, zhang2025agentracer, zhu2025raffles, west2025abduct, banerjee2025did, kong2025aegis}. For example, RAFFLES~\cite{zhu2025raffles} employs a Judge-Evaluator iterative framework to localize causal faults through multi-component collaboration, while ECHO~\cite{banerjee2025did} mitigates instability in long-trace analysis via hierarchical context modeling and multi-view voting. Complementary to these, a more specialized yet burgeoning paradigm shifts toward frameworks that decouple semantic extraction from diagnostic logic. For instance, FAMAS~\cite{ge2025introducing} performs statistical spectrum analysis across re-executed trajectories, while CDC-MAS~\cite{ma2025automatic} introduces a causal inference framework that identifies critical failure steps by modeling causal dependencies within execution trajectories.

Despite these advances, existing methods face significant challenges. LLM-based reasoning often incurs prohibitive computational costs and latency, while statistical approaches like spectrum analysis can struggle with contextual noise in complex trajectories. To address these issues, we propose StepFinder, a lightweight temporal semantic framework designed for efficient failure attribution. Unlike existing paradigms, StepFinder utilizes LLMs solely during the encoding stage to extract dense semantic representations. By delegating subsequent temporal modeling and relational reasoning to parameter-efficient deep learning modules, our approach achieves high-precision root cause localization with minimal computational overhead.

\section{Preliminaries}

\textbf{Multi-Agent System}. We focus on an LLM-based MAS $\mathcal{M}$, which comprises $N$ agents denoted by $\mathcal{N} = \{1, 2, \dots, N\}$. The system operates in discrete time steps, with agents acting in a round-robin manner, such that only one agent executes an action at each time step. The system can be formally described as:
\begin{equation} 
\mathcal{M} = (\mathcal{N}, \mathcal{S}, \mathcal{A}, \mathcal{P}, \phi)
\end{equation}
where $\mathcal{S}$ denotes the set of all possible system states, and $\mathcal{A}$ represents the set of all actions executable by the system. Each agent $i \in \mathcal{N}$ can only select actions from its corresponding action subset $\mathcal{A}_i \subseteq \mathcal{A}$. The function $\phi(t)$ identifies the agent that actually acts at time step $t$, thereby defining the round-robin execution order. The state transition probability $P(s_{t+1} \mid s_t, a_t, \phi(t))$ describes the likelihood of the system transitioning from state $s_t$ to $s_{t+1}$ when only agent $\phi(t)$ executes action $a_t$ at that time step. During execution, the system generates a record of states and actions, forming a complete trajectory denoted as $\tau = (s_0, a_0, s_1, a_1, \dots, s_T)$, where $T$ represents the final time step of the trajectory or the moment the system reaches a terminal state.

\noindent
\textbf{Root Cause Step}. We define an indicator function $Z(\tau) \in \{0,1\}$ to evaluate the task outcome, where $Z(\tau) = 1$ denotes a task failure and $Z(\tau) = 0$ denotes a successful task. To precisely define an error step, we introduce the notion of counterfactual intervention for any agent-time pair $(i, t)$ in trajectory $\tau$. If replacing the original action $a_t$ with a feasible corrective action $\tilde{a}_t$ while keeping the states prior to $t$ unchanged results in a modified trajectory $\tau^{(i,t)}$ that satisfies $Z(\tau^{(i,t)}) = 0$, then $(i, t)$ is considered a candidate error step.
Considering that long-range interactions may lead to error propagation, there may exist multiple candidate steps in a single trajectory. Guided by Occam’s Razor and the cascading failure mechanism, we focus on the earliest error step, which is defined as the decisive root cause of the task failure:
\begin{equation}
(i^\ast, t^\ast) = \arg\min_{(i,t)} \{ t \mid Z(\tau) = 1 , \ Z(\tau^{(i,t)}) = 0 \}  
\end{equation}

The objective of this work is, given a failed trajectory $\tau$, to automatically identify the decisive root cause step $t^\ast$ through deep modeling of trajectory behaviors.

\section{Methodology}

Our approach consists of three main components:

\textbf{(1) Trajectory Encoding.} We encode both the content and the agent of each step in MAS trajectories into embeddings, thereby constructing the entire execution trajectory as a temporal semantic sequence for subsequent model processing.

\textbf{(2) Model Architecture.} To capture both temporal dependencies and cross-step interactions, we design a network architecture that combines temporal modeling and agent-aware interaction modeling to learn representations of the temporal semantic sequences. For each step, an initial error score is computed, which is then adjusted using multi-scale differences and position bias to obtain the final anomaly score.

\textbf{(3) Loss Function.} The model is primarily trained with a supervised classification loss, supplemented by a self-supervised auxiliary loss that predicts the embedding of each step. This enhances trajectory pattern representation and improves the stability and generalization of the failure attribution model.

To further clarify our approach, the detailed algorithmic procedure is illustrated in Figure~\ref{fig:model_architecture} and formally presented as pseudocode in Appendix~\ref{appendix:Pseudocode}.

\begin{figure*}[t]
    \centering
    \includegraphics[width=\textwidth]{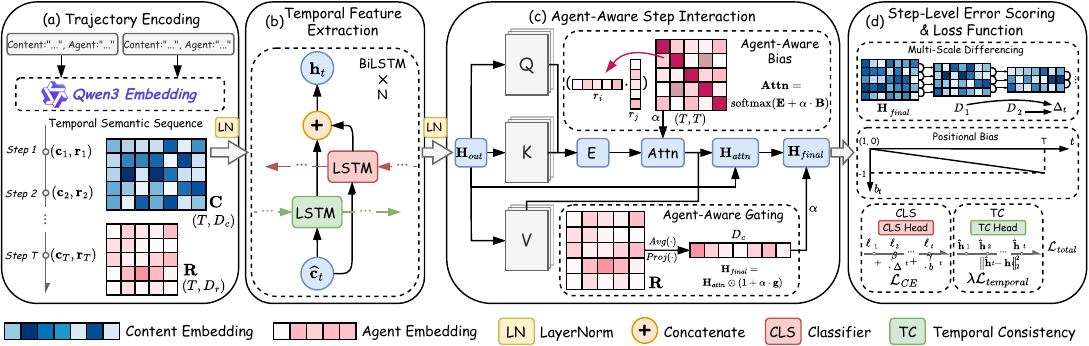} 
    \caption{Overview of StepFinder. Given a failure trajectory, execution logs are first encoded into temporal semantic sequences via Qwen3 Embedding (a). The content embeddings are then passed through the Temporal Feature Extraction module to capture temporal dependencies (b), followed by the Agent-Aware Step Interaction module that models cross-step dependencies with agent identity awareness (c). Finally, the Step-Level Error Scoring \& Loss Function module computes an explicit anomaly score for each step to identify the root cause, and optimizes the model via a joint loss (d).}
    \label{fig:model_architecture}
    \Description{Architecture.}
\end{figure*}

\subsection{Trajectory Encoding}

To accurately predict error steps, we first map the raw MAS trajectory into a high-dimensional vector space. As shown in Figure~\ref{fig:model_architecture}(a), each step $x_t=(c_t,r_t)$ consists of the execution content $c_t$ generated by the agent at step $t$ and the corresponding agent identity $r_t$. We employ Qwen3 Embedding~\cite{zhang2025qwen3} to produce embeddings $\mathbf{c}_t = \mathrm{Enc}(c_t) \in \mathbb{R}^{d_c}$ and $\mathbf{r}_t = \mathrm{Enc}(r_t) \in \mathbb{R}^{d_r}$, transforming the trajectory into two parallel feature matrices $\mathbf{C} \in \mathbb{R}^{T \times d_c}$ and $\mathbf{R} \in \mathbb{R}^{T \times d_r}$.

Under this formulation, the MAS execution is modeled as a semantic temporal sequence $\mathbf{X} = (\mathbf{C}, \mathbf{R})$. Here, content features $\mathbf{C}$ serve as the primary temporal signal, while agent features $\mathbf{R}$ provide auxiliary context to modulate step-specific behaviors, enabling a fine-grained analysis of anomalies and root causes at each step.

\subsection{Temporal Feature Extraction}

The step-level semantic trajectories generated by agent collaboration exhibit significant long-range dependencies and highly dynamic evolution, where small deviations in early steps may be amplified through frequent interactions among agents, resulting in cascading failure effects. To handle these characteristics, as shown in Figure~\ref{fig:model_architecture}(b), we design a Temporal Feature Extraction module aimed at capturing the bidirectional causal logic within the execution trajectory to accurately locate the root cause step.

First, we apply layer normalization~\cite{ba2016layer} to the content embedding sequence $\hat{\mathbf{C}} = \mathrm{LayerNorm}(\mathbf{C}) = \{\hat{\mathbf{c}}_1, \hat{\mathbf{c}}_2, \dots, \hat{\mathbf{c}}_T\}$. This operation mitigates magnitude differences across embedding dimensions and alleviates gradient instability during the training of long sequences, ensuring that the following model receives well-conditioned inputs across trajectories of varying lengths and complexity.
Subsequently, the normalized features are fed into a BiLSTM~\cite{hochreiter1997long} to capture bidirectional temporal dependencies. Specifically, the hidden state at each step is computed as $\mathbf{h}_t = [\overrightarrow{\text{LSTM}}(\hat{\mathbf{c}}_t); \overleftarrow{\text{LSTM}}(\hat{\mathbf{c}}_t)]$. The forward hidden state $\overrightarrow{\mathbf{h}}_t$ encodes the cumulative execution state from the beginning of the task to the current step, while the backward hidden state $\overleftarrow{\mathbf{h}}_t$ integrates potential evaluations from subsequent steps. This bidirectional modeling allows the identification of critical error steps that may appear reasonable locally but lead to task failure in the overall execution.
Finally, to ensure the robustness of the hidden state distribution and provide stable input features for subsequent relational modeling, we apply layer normalization again to the output hidden states $\mathbf{H} = \{ \mathbf{h}_t \}_{t=1}^T$, resulting in the final normalized representation $\mathbf{H}_{out} = \mathrm{LayerNorm}(\mathbf{H})$.

\subsection{Agent-Aware Step Interaction}

Although the Temporal Feature Extraction module effectively models local semantic evolution within execution trajectories, its chain-based propagation between adjacent states remains fundamentally limited to sequential dependencies. In multi-agent interaction scenarios, this limitation makes it difficult for the model to explicitly capture latent logical relationships between non-adjacent steps. Moreover, agent identity as a structural prior is not fully utilized, restricting the model’s ability to represent behavioral differences across agents. 

To address these limitations, as shown in Figure~\ref{fig:model_architecture}(c), we design the Agent-Aware Step Interaction module, in which each trajectory step interacts with all others through global attention, enabling the model to directly quantify semantic relationships between any pair of steps and capture deep, non-adjacent causal dependencies. Simultaneously, agent identity information is incorporated into the attention computation to modulate cross-step interactions, ensuring that the model accounts for behavioral differences when identifying critical step dependencies.

Specifically, we first map the node feature sequence output by the Temporal Feature Extraction module, $\mathbf{H}_{out} \in \mathbb{R}^{T \times d}$, into a multi-head attention space through linear projections with learnable weight matrices $W_Q$, $W_K$, and $W_V$ to generate the query $\mathbf{Q}$, key $\mathbf{K}$, and value $\mathbf{V}$ vectors.

To quantify the global semantic interaction between steps, the model computes the scaled dot-product interaction matrix $\mathbf{E}$ by correlating the query and key representations in each attention head:
\begin{equation}
\mathbf{E} = \frac{\mathbf{Q} \mathbf{K}^\top}{\sqrt{d_k}}
\end{equation}

\subsubsection{Agent-Aware Bias.} On this basis, to address the lack of agent identity awareness in the standard attention mechanism, we introduce an agent-aware bias matrix $\mathbf{B} \in \mathbb{R}^{T \times T}$, where $B_{ij} = \cos(\mathbf{r}_i, \mathbf{r}_j)$ represents the cosine similarity between the agent embeddings $\mathbf{r}_{i}$ and $\mathbf{r}_{j}$ corresponding to steps $i$ and $j$. We incorporate this structural prior as a bias term added to the semantic interaction scores, and use a coefficient $\alpha$ to control the strength of its influence. This guides the model to prioritize behavior chains with the same agent background, resulting in the final attention distribution:
\begin{equation}
\mathbf{Attn} = \text{softmax}(\mathbf{E} + \alpha \cdot \mathbf{B})    
\end{equation}

The model adaptively aggregates the feature information of each step according to the obtained attention weights, and employs a residual connection to ensure stable feature propagation, resulting in the attention-augmented features $\mathbf{H}_{attn} = \mathbf{H}_{out} + \mathbf{Attn} \cdot V$.

\subsubsection{Agent-Aware Gating.} Finally, to incorporate the global context of the multi-agent interaction trajectory, the model employs an agent-aware gating mechanism. We compute the global mean vector of the agent embeddings $\bar{\mathbf{r}} = \frac{1}{T} \sum_{t=1}^T \mathbf{r}_t$ to represent the collective agent profile. An auxiliary gating module then utilizes this global representation to generate modulation weights $\mathbf{g}$ for a second-order adjustment of the aggregated step features.
\begin{equation}
\mathbf{g} = \sigma(W_g \bar{\mathbf{r}} + b_g), \quad \mathbf{H}_{final} = \mathbf{H}_{attn} \odot (1 + \alpha \cdot \mathbf{g})
\end{equation}

\noindent
where $\sigma(\cdot)$ denotes the Sigmoid activation function. This gating mechanism dynamically adjusts the contribution of each step’s features based on global agent information, enabling the model to focus more on key information related to specific agent behavior chains during attribution. The resulting representations are ultimately used for the subsequent erroneous step localization task.

\subsection{Step-Level Error Scoring}

To quantify error probabilities, as shown in Figure~\ref{fig:model_architecture}(d), the model employs a lightweight nonlinear projection module that maps the step-level features to corresponding error scores. Specifically, the preliminary error score vector $\boldsymbol{\ell} \in \mathbb{R}^{T}$ is computed as:
\begin{equation}
\boldsymbol{\ell} = W_2 \cdot \operatorname{GELU}\left( W_1 \mathbf{H}_{final} + b_1 \right) + b_2
\end{equation}

These scores represent the model’s preliminary assessment of potential errors at each time step.

\subsubsection{Multi-Scale Differencing}

During multi-agent collaboration, errors often manifest as abnormal temporal fluctuations in latent representations rather than isolated semantic deviations. Such variations may emerge abruptly between adjacent steps or accumulate gradually over longer temporal spans. To capture dynamic state features at different temporal scales, we introduce a multi-scale temporal differencing mechanism on the top of the step-level representations. Specifically, for a given scale $s$, the temporal difference intensity at step $t$ is defined as:
\begin{equation}
D_t^{(s)} = \left\| \mathbf{h}_t - s \cdot \mathbf{h}_{t-1} + (s-1)\cdot \mathbf{h}_{t-s} \right\|_2    
\end{equation}

\noindent
when $s=1$, the above reduces to a first-order temporal difference, capturing local state transitions between adjacent steps; when $s>1$, it reflects the trend of changes over longer time spans, revealing the cumulative effect of erroneous behavior in mid-range temporal dynamics. To eliminate scale differences and highlight the relative strength of anomalies, the difference features at each scale are normalized as $\tilde{D}_t^{(s)}$, after which the model averages these normalized features across the multi-scale set $\mathcal{S}$ to obtain the final multi-scale temporal difference score $\Delta_t$:
\begin{equation}
\tilde{D}_t^{(s)} = \frac{D_t^{(s)}}{\frac{1}{T}\sum_{t=1}^{T} D_t^{(s)} + \epsilon}, \quad
\Delta_t = \frac{1}{|\mathcal{S}|} \sum_{s \in \mathcal{S}} \tilde{D}_t^{(s)}
\end{equation}

\subsubsection{Position Bias}

From an attribution perspective, errors in multi-agent execution trajectories are frequently rooted in critical early-stage decisions, while later anomalies tend to be downstream consequences of these initial mistakes~\cite{yao2022react}. However, discriminative models that rely solely on step-level representations tend to attribute errors to anomalous behaviors occurring at later stages of the trajectory. To mitigate this bias, we introduce an explicit position bias into step-level error prediction, imposing a structural constraint on the error likelihood across different time steps. For a trajectory of length $T$, the position bias at step $t$ is defined as:
\begin{equation}
b_t \;=\; - \frac{t-1}{T-1}, \quad t = 1, 2, \dots, T
\end{equation}

This linearly decaying bias serves as a mild structural prior favoring earlier steps. Semantic and temporal representations provide the primary signals for attribution. The position bias mainly helps resolve ambiguous cases where multiple steps exhibit similar error patterns, and its influence remains secondary to the learned representations when the true root cause occurs at later stages.

Combining preliminary error scores, multi-scale differences, and position bias, the final step-level error score is defined as:
\begin{equation}
\ell_t^{\mathrm{final}} = \ell_t + \beta \cdot \Delta_t + \gamma \cdot b_t
\end{equation}

\noindent
where $\beta$ and $\gamma$ are tunable hyperparameters that balance the influence of dynamic variation intensity and temporal structural bias in the attribution decision. The resulting score $\ell_t^{\mathrm{final}}$ is used to rank and localize potential erroneous steps within the execution trajectory.

\subsubsection{Temporal Consistency Head}

To explicitly model the temporal evolution of hidden states, we introduce a lightweight temporal prediction head within the attribution decision module, which employs a parameterized mapping function $f_{\theta}(\cdot)$ to project the hidden state at step $t$, denoted as $\mathbf{h}_t$, to the predicted representation of the next step $\hat{\mathbf{h}}_{t+1} = f_{\theta}(\mathbf{h}_t)$.

By incorporating this auxiliary prediction task, the model is encouraged to capture latent local temporal dependencies within multi-agent interaction trajectories, thereby imposing a temporal consistency constraint on the learned hidden representations. This design enables the model to better capture overall execution patterns while maintaining sensitivity to isolated erroneous steps.

\subsection{Loss Function}

To train the model to accurately discriminate the error occurrence probability at each step of a trajectory, we design a multi-component joint loss function. The overall objective is centered on standard classification supervision and is augmented with self-supervised constraints to enhance the model’s ability to capture latent logical anomalies in multi-agent interaction sequences.

\subsubsection{Classification Loss.}
The core optimization objective is to minimize the cross-entropy between the predicted step-level scores and the ground-truth error labels. Considering that execution trajectories may have variable lengths, padded steps are excluded from loss computation. Specifically, the cross-entropy loss is defined as:
\begin{equation}
\mathcal{L}_{\mathrm{CE}} = -\frac{1}{B} \sum_{i=1}^{B} \log \frac{\exp(\ell^{\mathrm{final}}_{i, y_i})} {\sum_{j \in \mathcal{T}_i} \exp(\ell^{\mathrm{final}}_{i, j})}
\end{equation}

\noindent
where $y_i$ denotes the ground-truth index of the faulty step for the $i$-th trajectory, and $\mathcal{T}_i$ represents the set of valid steps. This loss encourages the model to assign the highest confidence to the true error step within the global prediction distribution.

\subsubsection{Temporal Consistency Loss.}

In multi-agent interaction trajectories, faulty steps often mark a transition of the system’s hidden states from orderly collaboration to logical deviation. Traditional classification losses, which focus solely on isolated step-wise label matching, tend to overlook the coherent evolution of states along the temporal dimension, making it difficult for the model to distinguish transient semantic noise from true logical faults. To address this limitation, we introduce a temporal consistency loss $\mathcal{L}_{\mathrm{temporal}}$ as an auxiliary constraint:
\begin{equation}
\mathcal{L}_{\mathrm{temporal}} = \frac{1}{T-1} \sum_{t=2}^{T} \big\| \hat{\mathbf{h}}_t - \mathbf{h}_t \big\|_2^2
\end{equation}

\noindent
where $\hat{\mathbf{h}}_t$ is the predicted hidden representation of the current step based on the preceding state $\mathbf{h}_{t-1}$. This loss trains the model in a self-supervised manner to forecast future hidden states, enabling it to detect steps that deviate from the normal evolution pattern and thereby more sensitively capture potential errors that disrupt the coherence of multi-agent collaboration.

\subsubsection{Joint Training Objective}

The final training objective is realized through a weighted combination of the primary supervision loss and the auxiliary constraint:
\begin{equation}
\mathcal{L}_{\mathrm{total}} = \mathcal{L}_{\mathrm{CE}} + \lambda \mathcal{L}_{\mathrm{temporal}}
\end{equation}

\noindent
where $\lambda$ controls the contribution of the auxiliary constraint during training. This strategy enables the model to jointly optimize global classification accuracy and temporal logical consistency. The auxiliary constraint is applied only during the training phase; during testing and inference, the model retains only the classification branch and directly selects the step with the highest predicted score as the root cause for error localization.

\section{Experiments}

\subsection{Datasets}

We primarily evaluate our method on the Who\&When benchmark~\cite{zhang2025agent}, which serves as a standardized evaluation foundation for multi-agent failure attribution. This benchmark provides two distinct subsets of failure trajectories: Algorithm-Generated (Alg), which includes failures produced by automated systems, and Hand-Crafted (HC), which contains manually designed scenarios. Following the established protocol, we adopt the official test sets from both categories for evaluation, which consist of 126 trajectories for the Alg subset and 58 for the HC subset.

For model training, we first execute the MAS on the training tasks to obtain initial failure trajectories. To increase trajectory diversity, we further adopt a trajectory regeneration strategy that synthesizes new failed executions while preserving the original task semantics. Specifically, given an initial failure trajectory, we prompt an LLM to regenerate a different reasoning path and a different root cause error using the same question, ground truth answer, and agent team configuration. The root cause step and responsible agent are simultaneously generated during trajectory synthesis to provide fine grained supervision signals. Using this process, we generate 17 trajectories per task for the Alg subset across 92 training tasks and 14 trajectories per task for the HC subset across 186 training tasks, resulting in 1,564 and 2,604 training trajectories, respectively. The training and test sets are strictly disjoint at the task level to ensure reliable evaluation.

\subsection{Baselines}

To comprehensively evaluate the performance of the proposed model, we select the following three representative categories of baseline methods:

\textbf{(1) Random Attribution.}
Serving as a lower-bound reference, this method randomly samples an agent step from the interaction logs. It quantifies the inherent difficulty of the task by reflecting the baseline search space scale.

\textbf{(2) LLM-based Attribution.}
This category represents the current zero-shot attribution paradigm, where a pre-trained large language model is used to localize root cause steps in execution trajectories without domain-specific fine-tuning. Following the protocol in Who\&When, we evaluate three search strategies: \textbf{1) All-at-Once}. The LLM receives the full failure log and identifies the root cause step in a single inference. \textbf{2) Step-by-Step}. The LLM examines each step sequentially, terminating as soon as a mistake is identified and returning the corresponding step number. \textbf{3) Binary Search}. The LLM is asked whether the mistake lies in the upper or lower half of the log, and the selected half is recursively narrowed until a single step remains. Each method is evaluated both without access to the final ground truth of the task question to assess blind attribution capability, and with it ($\mathcal{G}$) to evaluate performance under additional information.

\textbf{(3) Sequential Models.} 
Since we formulate the execution trajectory as a temporal semantic sequence, we compare StepFinder against three classical sequence 
architectures representing the three dominant paradigms in sequential 
modeling: \textbf{1) BiGRU (RNN-based)} captures bidirectional temporal 
context through recurrent state propagation. \textbf{2) TCN (CNN-based)} employs dilated causal convolutions to capture long-range receptive 
fields with parallel computation. \textbf{3) Transformer (Attention-based)} models global dependencies across all steps via self-attention. These baselines evaluate whether generic sequential processing can match our specialized approach.

\subsection{Metrics}

Following established protocols in fault localization and failure attribution, we evaluate model performance across two dimensions: step-level attribution precision and the overall ranking quality of suspicious steps.

\textbf{(1) Attribution Precision.} We employ \textbf{Accuracy}~\cite{zhang2025agent} to check whether the step with the highest predicted score matches the ground truth. To account for temporal offsets, we also report \textbf{Tolerance Accuracy}~\cite{zhang2025agent}, which considers a prediction correct if it falls within a window of $\delta \in \{1, \dots, 5\}$ steps around the error source.

\textbf{(2) Ranking Quality.} We utilize two auxiliary metrics to assess ranking quality. \textbf{Accuracy@$K$ (Acc@$K$)}~\cite{ikram2022root} measures the probability of the root cause step appearing within the top-K candidates, reflecting retrieval coverage. \textbf{Mean Reciprocal Rank@3 (MRR@3)}~\cite{lin2024root} evaluates the average reciprocal rank of the true error among these candidates, representing the model's ability to prioritize the most likely root cause (formal definitions are provided in Appendix~\ref{appendix:MRR@3}).

\begin{table}[t]
    \centering
    \caption{Hyperparameter settings for Alg and HC subsets.}
    \label{tab:hyperparameters}
        \begin{tabular}{ccc}
            \toprule
            \textbf{Hyperparameter} & \textbf{Alg} & \textbf{HC} \\
            \midrule
            lr       & 1e-3   & 1e-5   \\
            $\alpha$ & 0.1    & 0.3    \\
            $\beta$  & 0.9    & 0.1    \\
            $\gamma$ & 0.4    & 0.75   \\
            $\lambda$& 0.9    & 0.02   \\
            $s$      & \{1, 2\} & \{1, 2\} \\
            \bottomrule
        \end{tabular}%
\end{table}

\subsection{Implementation Details}

All experiments are implemented based on Python 3.12, PyTorch 2.3.0, and CUDA 12.1, and all experiments are conducted on a single NVIDIA RTX 3080 Ti GPU. The specific configurations for each method are as follows:

\textbf{(1) LLM-based Attribution.} We use \texttt{\mdseries GPT-4o} following the official Who\&When~\cite{zhang2025agent} protocols for attribution precision metrics. For ranking quality, we refine the default prompt from the Who\&When benchmark to elicit the top-3 candidate error steps. Notably, only the All-at-Once strategy is used for ranking analysis, as its holistic output naturally aligns with ranking logic, unlike iterative approaches.

\textbf{(2) Sequential Models.} We replace the Temporal Feature Extraction and Agent-Aware Step Interaction modules in StepFinder with BiGRU, TCN, and Transformer, respectively, while keeping the same trajectory encoding module. We use the same projection head as StepFinder to compute step-level logits, followed by a standard cross-entropy loss for optimization. All sequential baselines are trained using a training pipeline identical to StepFinder's, with hidden dimensions and optimization settings kept consistent with it for fair comparison.

Specifically, we implement the BiGRU with 2 layers. The TCN baseline is configured with 2 layers using a kernel size of 3, where the dilation factor doubles per layer ($2^i$) to expand the receptive field. For Transformer, we employ 2 layers of encoder blocks with 4 attention heads and sinusoidal positional encodings.

\textbf{(3) StepFinder.} In the encoding stage, interaction contents are projected into dense vectors using \texttt{\mdseries Qwen3-Embedding-0.6B}. A timestep masking mechanism is integrated throughout the model to handle varying trajectory lengths, ensuring that attention computations are restricted to valid interactions, including masking invalid steps during loss computation. For model configuration, trajectory content and agent information are encoded into 128-dimensional and 32-dimensional vectors, respectively. The Temporal Feature Extraction Module employs a two-layer BiLSTM with a hidden dimension of 64 and a dropout rate of 0.5 to mitigate overfitting. In the Agent-Aware Step Interaction Module, the projection dimension for each attention head is set to 32, with 2 attention heads in total. The model is trained for up to 50 epochs with a batch size of 16 using the AdamW optimizer and a weight decay of 1e-5. Gradient clipping with a threshold of 1.0 and early stopping with a patience of 10 epochs are applied during training.

Considering the differences between the Alg and HC subsets, we keep the overall architecture fixed while tuning several key hyperparameters. The corresponding settings are summarized in Table~\ref{tab:hyperparameters}. All hyperparameters are selected through grid search.

\begin{table}[t]
    \centering
    \caption{Performance comparison on Who\&When benchmark. Accuracy is reported as mean $\pm$ std. dev. (\%). For LLM-based methods, $\mathcal{G}$ denotes those evaluated with ground truth, while others are without. The best results are in bold and second-best are \underline{underlined}.}
    \definecolor{graybg}{gray}{0.9} 
    \label{tab:main_results}
    \begin{tabular}{llcc}
        \toprule
        \textbf{Category} & \textbf{Method} & \textbf{Alg} & \textbf{HC} \\
        \midrule
        Baseline & Random & $15.08 \pm 1.71$ & $4.60 \pm 3.54$ \\
        \midrule
        \multirow{6}{*}{LLM-based} & All-at-Once & $11.90 \pm 1.59$ & $2.87 \pm 1.99$ \\
        & Step-by-Step & $14.02 \pm 1.21$ & $7.47 \pm 2.63$ \\
        & Binary Search & $15.61 \pm 1.21$ & $6.32 \pm 1.00$ \\
        & All-at-Once ($\mathcal{G}$) & $15.87 \pm 2.86$ & $4.60 \pm 2.64$ \\
        & Step-by-Step ($\mathcal{G}$) & $20.11 \pm 2.55$ & $6.90 \pm 1.73$ \\
        & Binary Search ($\mathcal{G}$) & $20.10 \pm 2.00$ & $5.75 \pm 1.00$ \\
        \midrule
        \multirow{4}{*}{Sequential} & BiGRU & $23.28 \pm 0.99$ & $\underline{12.64 \pm 2.15}$ \\
        & TCN & $24.34 \pm 0.75$ & $10.35 \pm 2.44$ \\
        & Transformer & $\underline{24.87 \pm 1.35}$ & $9.19 \pm 0.81$ \\
        & \cellcolor{graybg}StepFinder & \cellcolor{graybg}$\mathbf{29.63 \pm 1.50}$ & \cellcolor{graybg}$\mathbf{22.99 \pm 2.15}$ \\
        \bottomrule
    \end{tabular}
\end{table}

\begin{figure}[t]
    \centering
    \begin{subfigure}[b]{0.49\columnwidth}
        \centering
        \includegraphics[width=\textwidth]{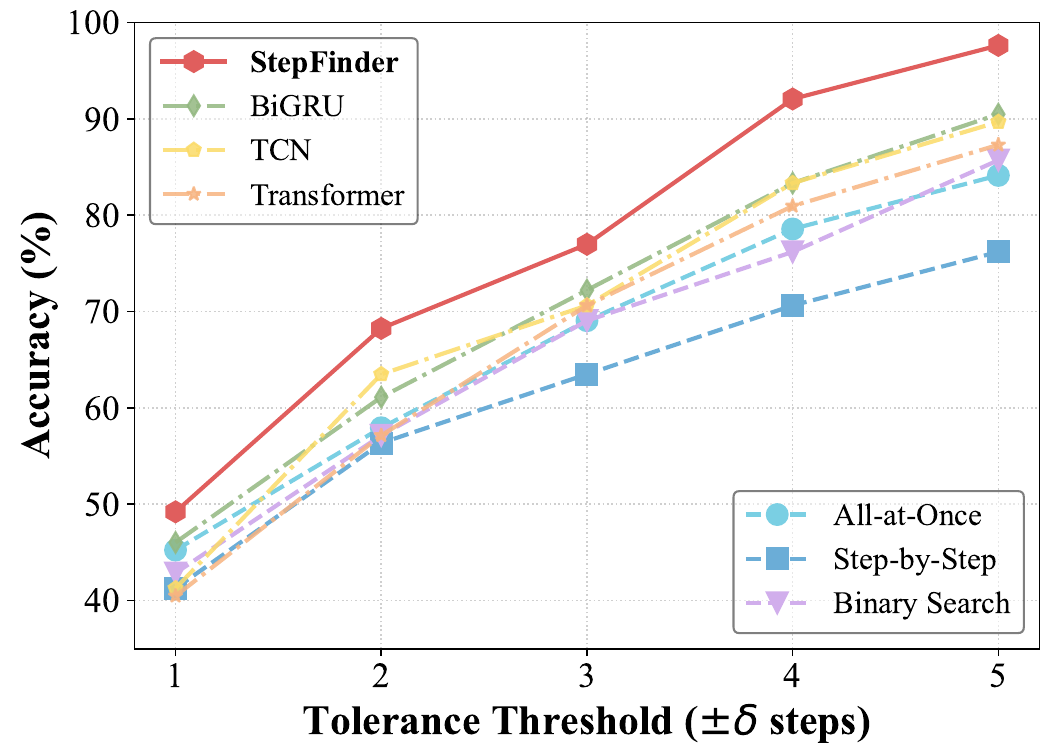}
        \caption{Alg subset}
        \label{fig:tolerance_algo}
    \end{subfigure}
    \hfill
    \begin{subfigure}[b]{0.49\columnwidth}
        \centering
        \includegraphics[width=\textwidth]{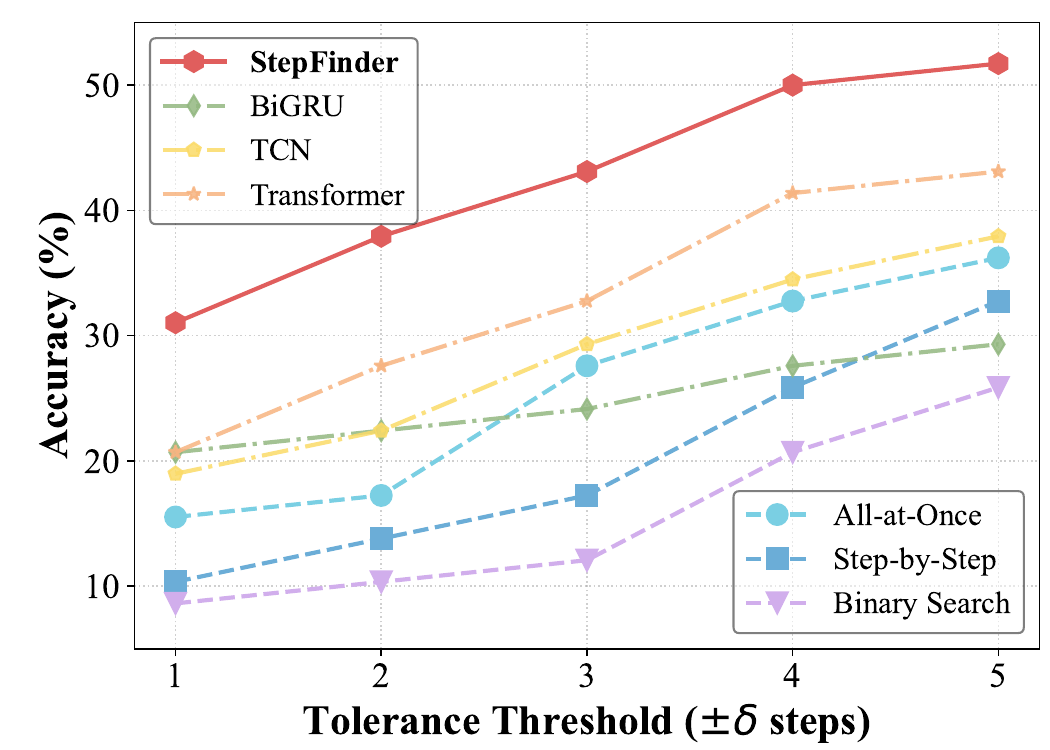}
        \caption{HC subset}
        \label{fig:tolerance_hand}
    \end{subfigure}
    
    \caption{Comparison of tolerance accuracy (\%) ($\pm \delta$ steps) on Who\&When.}
    \label{fig:tolerance_acc}
    \Description{Tolerance}
\end{figure}

\subsection{Main Results}

\subsubsection{Attribution Precision}
\textbf{(1) Accuracy.} The results in Table~\ref{tab:main_results} show that StepFinder consistently outperforms all baseline categories across both subsets.
\textbf{1) Comparison with Random Baseline}. 
StepFinder substantially outperforms Random attribution, demonstrating its capacity to capture diagnostic patterns from complex logs rather than relying on chance.
\textbf{2) Comparison with LLM-based Baselines}. 
StepFinder consistently outperforms LLM-based methods, surpassing the strongest competitor, Step-by-Step ($\mathcal{G}$), by 9.52\% on the Alg subset. On the more complex HC subset, it achieves an improvement of over 200\% compared to Step-by-Step. These results highlight the inherent limitations of LLM baselines in long-horizon causal reasoning for multi-agent trajectories.
\textbf{3) Comparison with Sequential Models}. 
Traditional sequential models generally outperform LLM-based methods, validating our use of dedicated sequential architectures for step-level fault attribution. However, StepFinder maintains a significant lead, surpassing the best baseline, BiGRU, by 10.35\% on the HC subset. This superiority stems from StepFinder's ability to integrate robust temporal dependencies with global inter-step interactions for precise root cause localization.
For completeness, we further compare with recent concurrent work in Appendix~\ref{appendix:concurrent}.

\textbf{(2) Tolerance Accuracy.} 
As shown in Figure~\ref{fig:tolerance_acc}, StepFinder maintains a dominant lead across both subsets. On the Alg subset, accuracy increases rapidly with the tolerance threshold $\delta$, primarily because the shorter trajectory lengths make each step increment more impactful. StepFinder outperforms the best LLM-based method by approximately 15\% at $\delta=1$, indicating that the predictions of StepFinder are highly concentrated around the actual root cause step. Beyond the optimized architecture, this precision stems from our multi-scale differencing and position bias, which capture fine-grained anomalies and prioritize early-stage errors.

\begin{table}[t]
\centering
\caption{Comparison of Acc@$K$ (\%) on Who\&When. The best results are in \textbf{bold}.}
\definecolor{graybg}{gray}{0.9} 
\label{tab:acc_k_results}
\begin{tabular}{clccc}
\toprule
\textbf{Dataset} & \textbf{Method} & \textbf{Acc@1} & \textbf{Acc@2} & \textbf{Acc@3} \\ \midrule
\multirow{5}{*}{\begin{tabular}[c]{@{}l@{}}Alg\end{tabular}} 
 & All-at-Once   & 15.87 & 22.22 & 28.57 \\
 & BiGRU         & 20.63 & 40.48 & 54.76 \\
 & TCN           & 23.81 & 42.86 & 52.38 \\
 & Transformer   & 25.40 & 41.27 & 56.35 \\
 & \cellcolor{graybg}StepFinder & \cellcolor{graybg}\textbf{28.84} & \cellcolor{graybg}\textbf{45.24} & \cellcolor{graybg}\textbf{60.31} \\ \midrule
\multirow{5}{*}{\begin{tabular}[c]{@{}l@{}}HC\end{tabular}} 
 & All-at-Once   & 12.07 & 15.52 & 18.97 \\
 & BiGRU         & 15.52 & 17.24 & 17.24 \\
 & TCN           & 12.07 & 15.52 & 18.97 \\
 & Transformer   & 10.34 & 17.24 & 22.41 \\
 & \cellcolor{graybg}StepFinder & \cellcolor{graybg}\textbf{21.26} & \cellcolor{graybg}\textbf{25.29} & \cellcolor{graybg}\textbf{30.46} \\ \bottomrule
\end{tabular}
\end{table}

\begin{figure}[t]
    \centering
    \includegraphics[width=0.9\columnwidth]{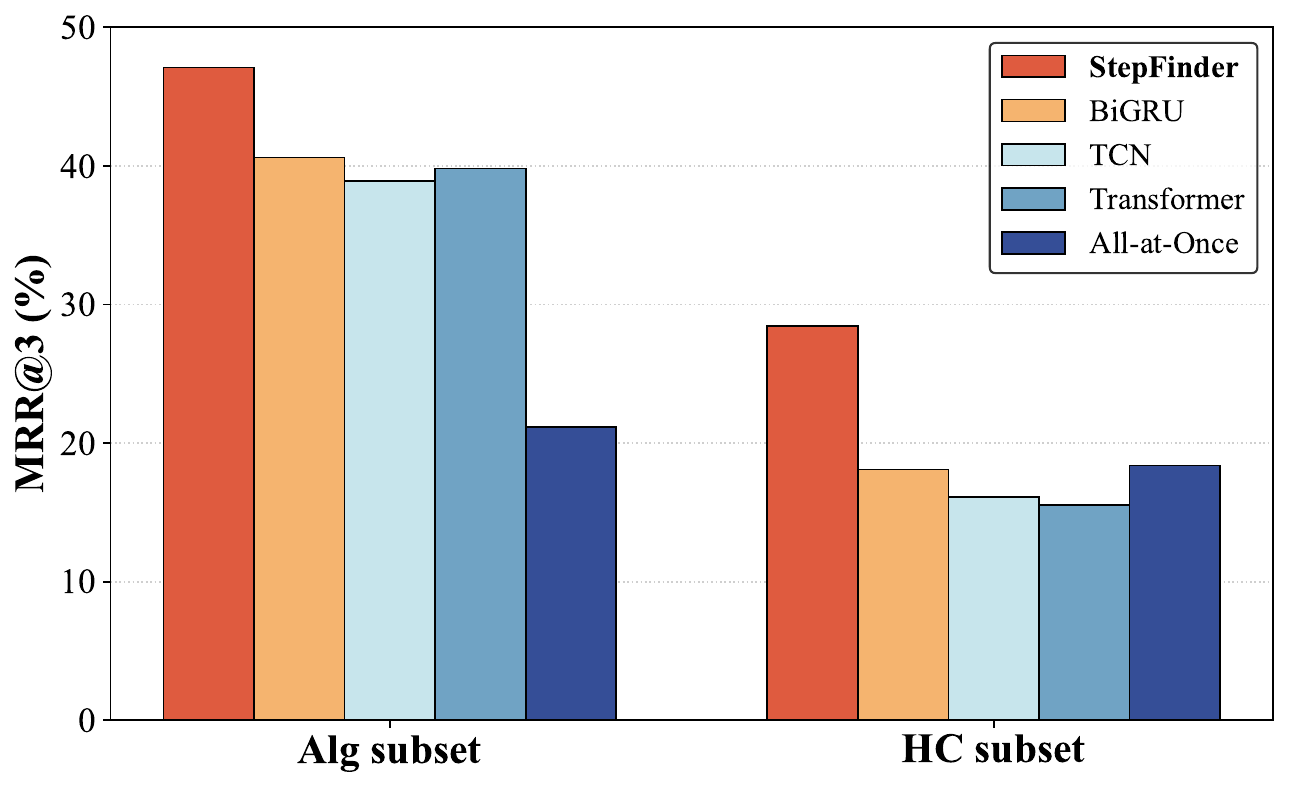}
    \caption{Comparison of MRR@3 (\%) on Who\&When.}
    \label{fig:mrr_results}
    \Description{MRR@3}
\end{figure}

\subsubsection{Ranking Quality}
\textbf{(1) Acc@$K$.}
For all sequential models, including StepFinder, we report the configuration achieving the best balanced performance across all metrics. As shown in Table~\ref{tab:acc_k_results}, StepFinder consistently outperforms all baselines in Acc@$K$ metrics across both subsets. On the Alg subset, StepFinder achieves a significant Acc@3 of 60.31\%, surpassing Transformer by nearly 4\%. More notably, on the HC subset, StepFinder's Acc@1 already exceeds the Acc@3 of most baselines. In contrast, the LLM-based method exhibits the weakest performance, with its Acc@3 on Alg failing to even match StepFinder's Acc@1. This highlights that direct LLM prompting lacks the necessary ranking quality for dense attribution, whereas StepFinder effectively prioritizes true root causes even in complex, long-range trajectories. 

Based on the ranked outputs of the model, we further evaluated its attribution performance under different Acc@$K$ candidate ranges, as shown in Table~\ref{tab:ranked_attribution}. On the Alg subset, when the search range expanded from Acc@1 to Acc@3, the attribution coverage increased significantly from 29.63\% to 64.81\%, indicating that more than 60\% of failures could be constrained within a very small set of candidates. For the more structurally complex and interaction-rich HC subset, although precise localization is more challenging, the model still achieved a maximum coverage of 35.06\% at Acc@3. These results demonstrate that ranking-based attribution can effectively mitigate instability caused by single-step decisions in long execution trajectories. In practical applications, by restricting potential faulty steps to a small set of high-risk candidates, the model substantially reduces the search space for human auditing, thereby improving the usability of MAS failure diagnosis.

\begin{table}[t]
\centering
\caption{StepFinder Acc@$K$ (\%) results under different selection criteria. Within each subset block, the three rows report results for models optimized for Acc@1, Acc@2, and Acc@3 respectively, with the targeted metric highlighted in bold.}
\label{tab:ranked_attribution}
\begin{tabular}{cccc}
\toprule
\textbf{Dataset} & \textbf{Acc@1} & \textbf{Acc@2} & \textbf{Acc@3} \\
\midrule
\multirow{3}{*}{Alg} & \textbf{29.63} & 43.65 & 60.84 \\
                                   & 29.10 & \textbf{48.41} & 57.94 \\
                                   & 26.98 & 46.83 & \textbf{64.81} \\
\midrule
\multirow{3}{*}{HC}        & \textbf{22.99} & 26.44 & 29.89 \\
                                   & 20.69 & \textbf{28.16} & 31.03 \\
                                   & 16.67 & 25.29 & \textbf{35.06} \\
\bottomrule
\end{tabular}
\end{table}

\textbf{(2) MRR@3.}
As shown in Figure~\ref{fig:mrr_results}, StepFinder leads in MRR@3 on both subsets, reaching nearly 50\% on the Alg subset. This performance outperforms the LLM-based method by over 25\% and BiGRU by more than 6\%. On the HC subset, although the LLM-based method marginally exceeds other sequential models, it still trails StepFinder by roughly 10\%. This result indicates that StepFinder not only includes the true error among its candidates but also ranks it at the top with high confidence, substantially reducing manual verification effort in complex multi-agent diagnostics.

\begin{table}[t]
    \centering
    \caption{Ablation study on Who\&When (Accuracy \%). The upper section compares alternative temporal encoders, and the lower section evaluates the contribution of each component by removing them individually. The best results are in bold.}
    \label{tab:ablation_study}
        \begin{tabular}{l l cc}
            \toprule
            \textbf{Category} & \textbf{Variant} & \textbf{Alg} & \textbf{HC} \\
            \midrule
            \multirow{3}{*}{Temporal Encoder} 
                & LSTM        & 25.93 & 9.20 \\
                & BiGRU       & 26.19 & 17.82 \\
                & Transformer & 26.45 & 16.67 \\
            \midrule
            \multirow{3}{*}{Architecture} 
                & w/o TFE   & 25.13 & 12.64 \\
                & w/o ASI   & 27.51 & 18.97 \\
                & w/o AI    & 28.04 & 19.54 \\
            \midrule
            \multirow{2}{*}{Error Score} 
                & w/o MsDiff & 29.37 & 19.54 \\
                & w/o PB     & 26.98 & 20.11 \\
            \midrule
            Loss Function 
                & w/o TCLoss & 25.13 & 16.67 \\
            \midrule
            StepFinder & --- & \textbf{29.63} & \textbf{22.99} \\
            \bottomrule
        \end{tabular}
\end{table}

\section{Ablation Study}

To analyze the impact of different design choices on model performance, we conduct ablation experiments on the Who\&When benchmark, examining both the selection of temporal encoder and the individual contribution of each component in StepFinder, with results summarized in Table~\ref{tab:ablation_study}.

We first validate the choice of BiLSTM as the temporal encoder in Temporal Feature Extraction (TFE) by comparing it with three representative sequence modeling methods. LSTM models information in a unidirectional manner and lacks access to future context, while BiGRU incorporates bidirectional information but remains limited in modeling long-range dependencies. Transformer relies on global attention without explicit sequential inductive bias, making it less effective under limited data scale and the strongly sequential structure of MAS trajectories. Experimental results show that BiLSTM consistently achieves the best performance among all variants.

Regarding the model architecture, we separately remove TFE, Agent-Aware Step Interaction (ASI), and Agent Identity (AI) modules. Here, AI refers collectively to the agent-aware bias and the agent-aware gating mechanisms. Removing TFE causes the largest performance drop, especially on the HC subset, highlighting the importance of temporal modeling for long-range step dependencies. Eliminating ASI also degrades performance, confirming the benefit of explicit step-level interaction modeling, though to a lesser extent. Furthermore, removing AI results in a smaller but consistent decline, indicating its role in improving behavioral consistency.

In terms of error score, Multi-scale Differencing (MsDiff) and Position Bias (PB) exhibit notable scenario-dependent effects. Removing MsDiff had minimal impact on the Alg subset but caused a clear performance drop on the HC subset, suggesting that it helps capture fine-grained anomalies in complex interaction patterns. In contrast, removing PB mainly affects the Alg subset, indicating that step position information aids in aligning execution flows in tasks with more regular logical structures.

At the level of the loss function, removing the Temporal Consistency Loss (TCLoss) substantially degrades accuracy, demonstrating the critical role of future-step prediction as a self-supervised signal in guiding the model to learn temporal consistency.

Overall, StepFinder achieves the best performance across both subset types, demonstrating the rationality of the overall design and the effectiveness of its individual components. Detailed sensitivity analysis of key hyperparameters is provided in Appendix~\ref{appendix:sensitivity}.

\section{Efficiency Analysis}

To evaluate the computational cost and inference efficiency of different methods in step-level fault attribution for multi-agent trajectories, we conduct an efficiency analysis. The analysis metrics include input tokens, output tokens, and processing time per sample. For StepFinder, the input and output tokens correspond to those used in the encoding stage, while the per-sample processing time includes both the encoding time and the model inference time.

As shown in Table~\ref{tab:token_time_comparison}, StepFinder demonstrates clear efficiency and low computational cost across both subsets. Specifically, in terms of input tokens, StepFinder is comparable to the All-at-Once method but significantly lower than Step-by-Step and Binary Search. Notably, StepFinder incurs zero generation tokens, as it bypasses text generation in favor of direct representation-based attribution. Regarding processing time, StepFinder requires only 0.61 seconds per sample on the Alg subset and 3.56 seconds per sample on the HC subset, saving roughly an order of magnitude compared to the Step-by-Step method. Overall, these results demonstrate that StepFinder can maintain accurate root cause localization while achieving highly efficient inference at minimal computational cost, making it well-suited for large-scale multi-agent trajectory analysis.

\begin{table}[t]
\centering
\caption{Efficiency comparison of token usage and inference time per sample on Who\&When. The minimum values are in bold.}
\label{tab:token_time_comparison}
\definecolor{graybg}{gray}{0.9}
\resizebox{\columnwidth}{!}{%
\begin{tabular}{clrrr}
\toprule
\textbf{Dataset} & \textbf{Method} & \textbf{Input Tokens} & \textbf{Output Tokens} & \textbf{Time (s)} \\
\midrule
\multirow{4}{*}{Alg} 
                    & All-at-Once      & 3,334.85   & 138.23  & 2.92 \\
                    & Step-by-Step     & 15,673.79  & 1,626.20 & 26.18 \\
                    & Binary Search    & 6,703.04    & 19.02  & 5.41 \\
                    & \cellcolor{graybg}StepFinder & \cellcolor{graybg}\textbf{3,289.35} & \cellcolor{graybg}\textbf{0.00} & \cellcolor{graybg}\textbf{0.61} \\
\midrule
\multirow{4}{*}{HC}        
                    & All-at-Once      & \textbf{17,025.60}  & 130.47 & 3.92 \\
                    & Step-by-Step     & 79,817.71  & 1,515.95 & 35.49 \\
                    & Binary Search    & 34,771.07  & 16.57 & 6.92 \\
                    & \cellcolor{graybg}StepFinder & \cellcolor{graybg}19,257.17  & \cellcolor{graybg}\textbf{0.00} & \cellcolor{graybg}\textbf{3.56} \\
\bottomrule
\end{tabular}
}
\end{table}

\section{Conclusion}

In this study, we propose StepFinder, an efficient step-level failure attribution method. The model uses LLMs only during the encoding stage for text representation learning and jointly models the temporal evolution and cross-step dependencies of execution steps to quantify the anomaly score of each step. StepFinder incorporates a multi-scale differencing mechanism to capture anomalous fluctuations, leverages step position information to focus more on early key steps, and uses a future-step prediction loss to guide temporal consistency learning, thereby enhancing its ability to identify root cause steps. Extensive experiments demonstrate that StepFinder significantly outperforms existing LLM-based attribution methods on the Who\&When benchmark, achieving accurate step-level failure attribution with substantially lower inference latency and computational cost.

\begin{acks}
This work is supported by the Postdoctoral Fellowship Program of CPSF under Grant Number GZC20251085 and the China Postdoctoral Science Foundation under Grant Number 2025M781445.
\end{acks}

\clearpage

\bibliographystyle{ACM-Reference-Format}
\bibliography{main}

\appendix

\section{Pseudocode of StepFinder}
\label{appendix:Pseudocode}

\begin{algorithm}[H]
\caption{StepFinder Training}
\begin{algorithmic}[1]
\REQUIRE MAS failure logs $\mathcal{T} = \{\tau_1, \tau_2, \dots, \tau_N\}$
\ENSURE Trained StepFinder model parameters $\Theta$

\FOR{$\tau$ in $\mathcal{T}$}
    \STATE \textcolor{blue}{\textit{/* Trajectory Encoding */}}
    \STATE Parse $\tau$ into content and agent features: $\{(c_t, r_t)\}_{t=1}^T$
    \STATE Encode content and agent using Qwen3 Embedding:
    \STATE \hspace{2em} $\mathbf{C} \leftarrow \{\mathrm{Enc}(c_t)\}_{t=1}^T,\;\mathbf{R} \leftarrow \{\mathrm{Enc}(r_t)\}_{t=1}^T$
    \STATE Construct step-level semantic trajectory $\mathbf{X} = (\mathbf{C}, \mathbf{R})$

    \STATE \textcolor{blue}{\textit{/* Temporal Feature Extraction */}}
    \STATE $\mathbf{H}_{out} \leftarrow \mathrm{LayerNorm}(\mathrm{BiLSTM}(\mathrm{LayerNorm}(\mathbf{C})))$

    \STATE \textcolor{blue}{\textit{/* Agent-Aware Step Interaction */}}
    \STATE $\mathbf{Q}, \mathbf{K}, \mathbf{V} \leftarrow \mathbf{H}_{out} W_Q, \mathbf{H}_{out} W_K, \mathbf{H}_{out} W_V$ 
    \STATE $\mathbf{E} \leftarrow \frac{\mathbf{Q} \mathbf{K}^\top}{\sqrt{d_k}}$ 
    \textcolor{gray}{// interaction score}
    \STATE $B_{ij} \leftarrow \cos(\mathbf{r}_i, \mathbf{r}_j)$
    \textcolor{gray}{// agent-aware bias matrix}
    \STATE $\mathbf{Attn} \leftarrow \text{softmax}(\mathbf{E} + \alpha \cdot \mathbf{B})$
    \STATE $\mathbf{H}_{attn} \leftarrow \mathbf{H}_{out} + \text{Attention}(\mathrm{Attn}, V)$
    \textcolor{gray}{// residual connection}
    \STATE $\mathbf{g} \leftarrow \sigma\Big(W_g \big(\frac{1}{T} \sum_{t=1}^T \mathbf{r}_t\big) + b_g\Big)$
    \textcolor{gray}{// modulation weights}
    \STATE $\mathbf{H}_{final} \leftarrow \mathbf{H}_{attn} \odot (1 + \alpha \cdot \mathbf{g})$
    \textcolor{gray}{// gating}

    \STATE \textcolor{blue}{\textit{/* Step-Level Error Scoring */}}
    \STATE $\boldsymbol{\ell} \leftarrow W_2 \cdot \mathrm{GELU}(W_1 \mathbf{H}_{final} + b_1)$
    \textcolor{gray}{// initial error scores}
    \FOR{$s \in \mathcal{S}$}
        \STATE \textcolor{gray}{// multi-scale differences}
        \STATE $D^{(s)}_t \leftarrow \|\mathbf{h}_t - s \cdot \mathbf{h}_{t-1} + (s-1) \cdot \mathbf{h}_{t-s}\|_2$
        \STATE $\tilde{D}^{(s)}_t \leftarrow D^{(s)}_t / (\frac{1}{T} \sum_{t=1}^T D^{(s)}_t + \epsilon)$
    \ENDFOR
    \STATE $\Delta_t \leftarrow \frac{1}{|\mathcal{S}|} \sum_{s \in \mathcal{S}} \tilde{D}^{(s)}_t$
    \STATE $b_t \leftarrow -\frac{t-1}{T-1}$
    \textcolor{gray}{// position bias}
    \STATE $\ell_t^{\mathrm{final}} \leftarrow \ell_t + \beta \cdot \Delta_t + \gamma \cdot b_t$
    \STATE $\hat{\mathbf{h}}_{t+1} \leftarrow f_\theta(\mathbf{h}_t)$
    \textcolor{gray}{// temporal consistency head}

    \STATE \textcolor{blue}{\textit{/* Loss Function */}}
    \STATE $\mathcal{L}_{\mathrm{CE}} \leftarrow \text{CrossEntropy}(\ell^{\mathrm{final}}, y)$
    \textcolor{gray}{// y: label}
    \STATE $\mathcal{L}_{\mathrm{temporal}} \leftarrow \frac{1}{T-1} \sum_{t=2}^{T} \|\hat{\mathbf{h}}_t - \mathbf{h}_t\|_2^2$
    \STATE $\mathcal{L}_{\mathrm{total}} \leftarrow \mathcal{L}_{\mathrm{CE}} + \lambda \cdot \mathcal{L}_{\mathrm{temporal}}$
    \STATE $\Theta \leftarrow \Theta - \eta \nabla_{\Theta} \mathcal{L}_{\mathrm{total}}$
\ENDFOR

\end{algorithmic}
\end{algorithm}

\begin{figure*}[t]
    \centering
    \begin{subfigure}{0.24\textwidth}
        \centering
        \includegraphics[width=\linewidth]{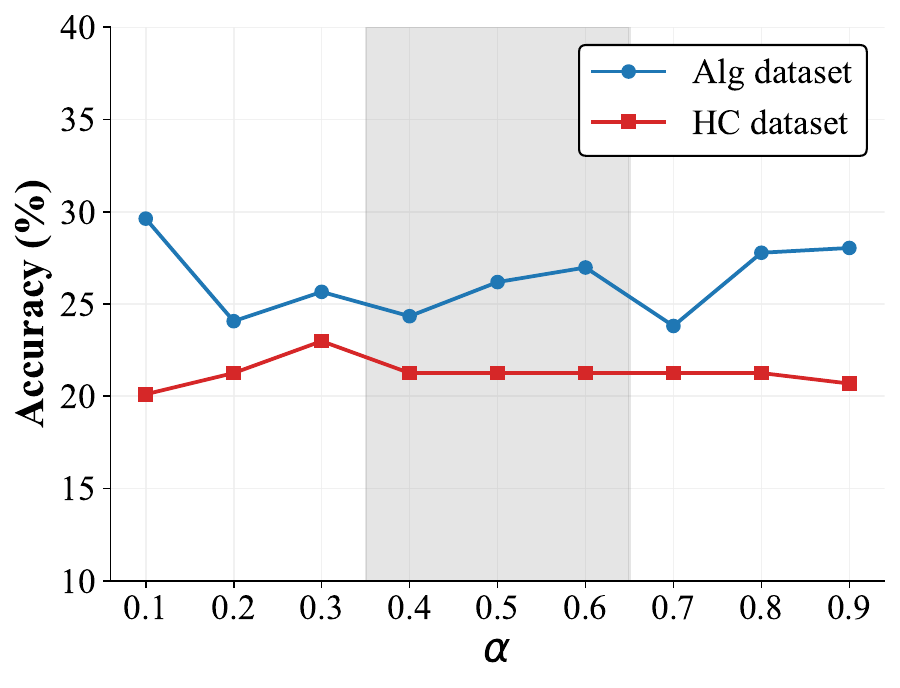}
        \caption{Sensitivity of $\alpha$}
        \label{fig:sens_alpha}
    \end{subfigure}
    \hfill
    \begin{subfigure}{0.24\textwidth}
        \centering
        \includegraphics[width=\linewidth]{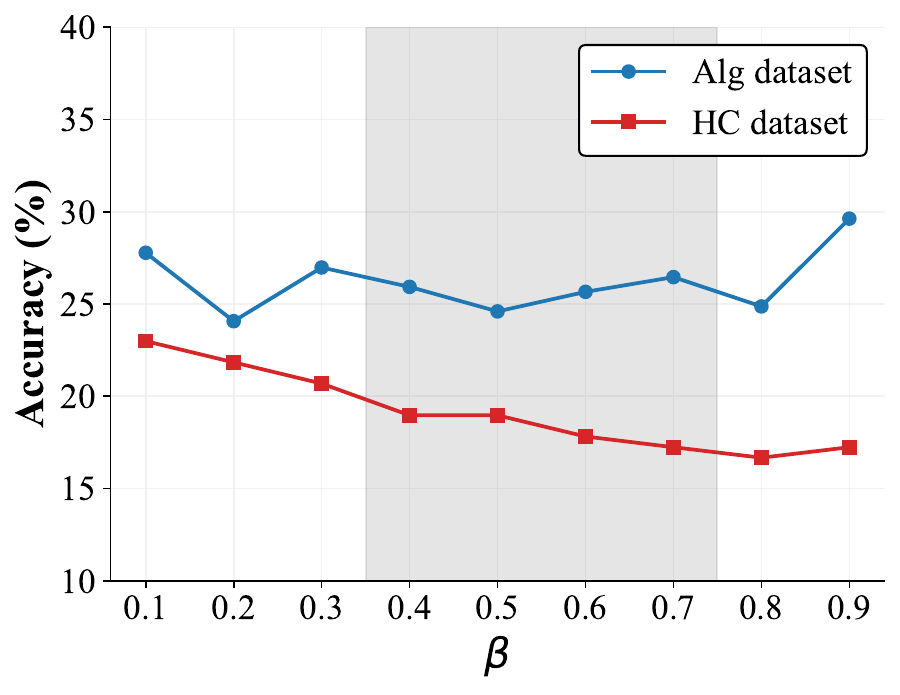}
        \caption{Sensitivity of $\beta$}
        \label{fig:sens_beta}
    \end{subfigure}
    \hfill
    \begin{subfigure}{0.24\textwidth}
        \centering
        \includegraphics[width=\linewidth]{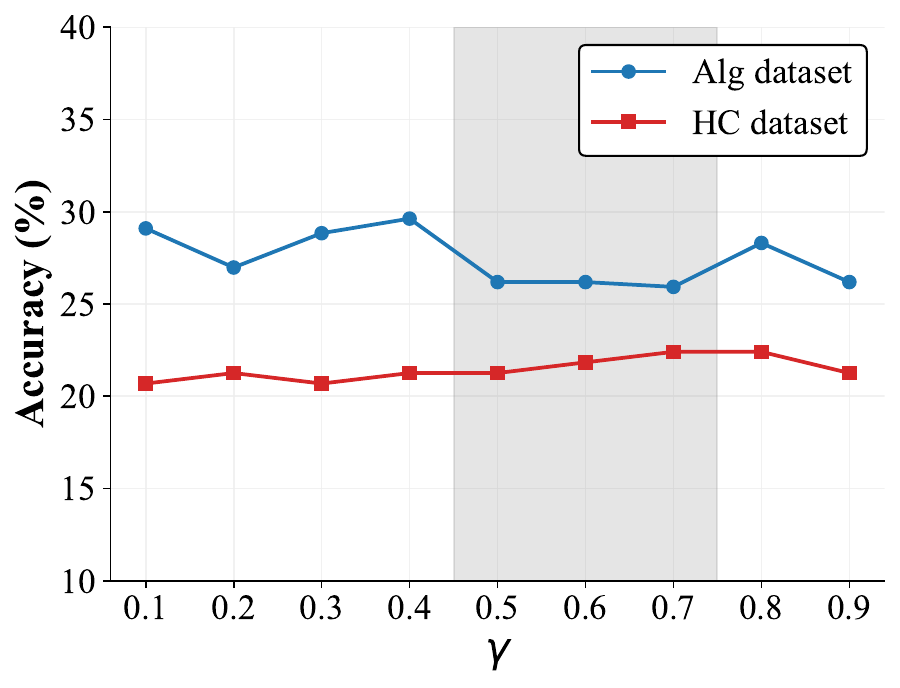}
        \caption{Sensitivity of $\gamma$}
        \label{fig:sens_gamma}
    \end{subfigure}
    \hfill
    \begin{subfigure}{0.24\textwidth}
        \centering
        \includegraphics[width=\linewidth]{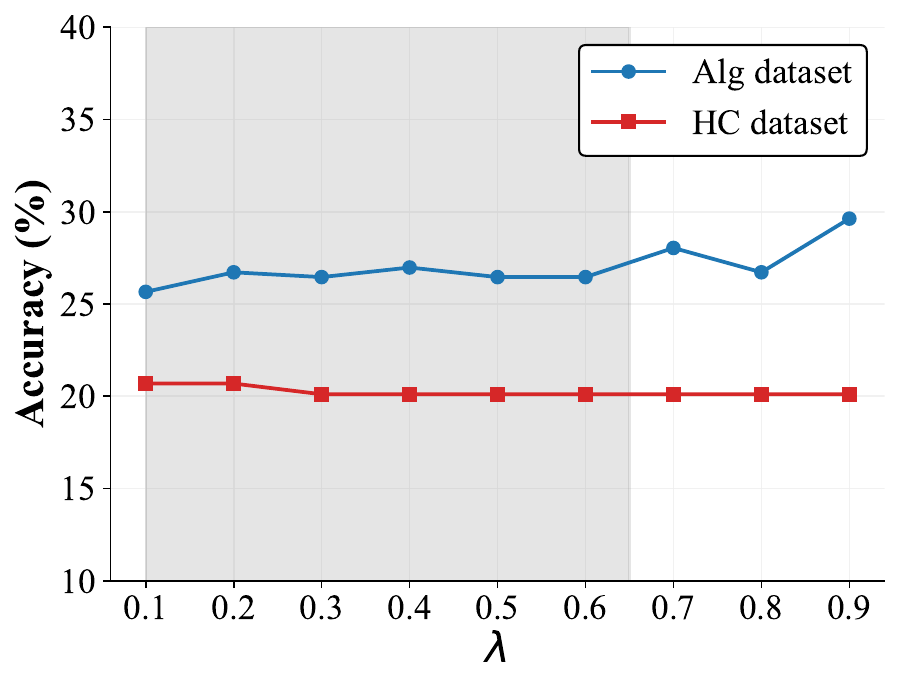}
        \caption{Sensitivity of $\lambda$}
        \label{fig:sens_lambda}
    \end{subfigure}

    \caption{Sensitivity analysis of key hyperparameters on Who\&When (Accuracy \%). The gray shaded areas indicate the robust regions where StepFinder maintains stable performance across both subsets.}
    \label{fig:overall_sensitivity}
    \Description{Sensitivity analysis}
\end{figure*}

\section{Mean Reciprocal Rank@3}
\label{appendix:MRR@3}
Mean Reciprocal Rank (MRR) is a standard statistic for evaluating models that produce a ranked list of potential causes. For a set of $N$ trajectories, it is calculated as the average of the reciprocal ranks of the ground-truth steps:
\begin{equation}
\text{MRR} = \frac{1}{N} \sum_{i=1}^{N} \frac{1}{\text{rank}_i}
\end{equation}
where $\text{rank}_i$ denotes the position of the true error step in the $i$-th predicted ranking. To better reflect the model performance in high-precision scenarios, we utilize MRR@$K$, which only considers the ground truth if it appears within the top-$K$ candidates. The reciprocal rank for a single instance is truncated as follows:
\begin{equation}
\text{MRR@}K =
\begin{cases}
\frac{1}{\mathrm{rank}} & \text{if } \mathrm{rank} \le K, \\
0                       & \text{otherwise}.
\end{cases}
\end{equation}

To align with the practical requirements of MAS maintenance, where diagnostic tools are expected to provide a highly concise set of candidate steps, we set $K=3$ in our experiments.

\section{Comparison with Concurrent Work}
\label{appendix:concurrent}
We further compare StepFinder with two recent failure attribution approaches: AgenTracer~\cite{zhang2025agentracer}, an LLM-based attribution method that directly performs trajectory reasoning, and CDC-MAS~\cite{ma2025automatic}, a causal inference framework that decouples semantic extraction from diagnostic analysis. Since both methods are not publicly available, we directly report the attribution accuracy values from their original papers and restrict the comparison to the shared metric.

As shown in Table~\ref{tab:concurrent}, StepFinder outperforms both methods on the HC subset, while achieving lower accuracy on Alg. The better performance of AgenTracer and CDC-MAS on the Alg subset can be attributed to the shorter and more structured trajectories, where LLM-based reasoning and causal analysis are more effective. On the HC subset, which features longer and more complex interaction patterns, LLM-based methods are more susceptible to interference from redundant context, and causal dependencies become harder to capture. The explicit temporal modeling and cross-step interaction in StepFinder provide stronger robustness under such conditions.

Regarding computational efficiency, AgenTracer relies on reinforcement learning to fine-tune an LLM, resulting in heavy training and inference overhead. CDC-MAS performs iterative multi-stage causal analysis over execution trajectories, which introduces additional inference overhead due to repeated reasoning steps. In contrast, StepFinder uses LLMs only for feature encoding and performs direct neural inference using a lightweight model, which leads to lower computational overhead.

\begin{table}[t]
    \centering
    \caption{Comparison with concurrent attribution methods on Who\&When (Accuracy \%). For AgenTracer, $\mathcal{G}$ indicates evaluation with ground truth. The best results are in bold.}
    \label{tab:concurrent}
    \begin{tabular}{lcc}
        \toprule
        \textbf{Method} & \textbf{Alg} & \textbf{HC} \\
        \midrule
        AgenTracer & 37.30 & 20.68 \\
        AgenTracer ($\mathcal{G}$) & \textbf{42.86} & 20.68 \\
        CDC-MAS & 36.20 & 18.20 \\
        StepFinder & 29.63 & \textbf{22.99} \\
        \bottomrule
    \end{tabular}
\end{table}

\section{Sensitivity Analysis}
\label{appendix:sensitivity}

To further evaluate the robustness and stability of the model, as shown in Figure~\ref{fig:overall_sensitivity}, we analyze the sensitivity of StepFinder to four key hyperparameters, including the agent-aware weight $\alpha$, multi-scale difference weight $\beta$, position bias weight $\gamma$, and temporal consistency weight $\lambda$. Each parameter is varied from 0.1 to 0.9 with a step size of 0.1 on both the Alg and HC subsets.

$\bm{\alpha}$ has a significant impact on model performance. Its effect is particularly noticeable on the Alg subset, where both low and high values achieve relatively strong results while medium values lead to fluctuations. This indicates that in well-structured systems, Agent Identity can serve either as a strong prior guiding the model or as a data-driven signal, whereas medium weights may cause competition between the prior and the signal. On the HC subset, performance remains stable and peaks at 0.3, suggesting that Agent Identity mainly provides robust auxiliary guidance rather than acting as a decisive factor in complex scenarios.

$\bm{\beta}$ exhibits different effects across the two subsets. On the Alg subset, model performance generally increases as $\beta$ grows, reaching a peak at 0.9, indicating that stronger difference constraints help extract stable signals from structured execution logs. In contrast, on the HC subset, performance decreases as $\beta$ increases, suggesting that overly strong smoothing constraints may hinder the model's ability to capture complex interactions.

$\bm{\gamma}$ yields distinct accuracy trends on both subsets. On the Alg subset, a moderate weight achieves the best performance, indicating that an appropriate position bias improves structural modeling between steps. On the HC subset, performance gradually increases with higher weights and peaks in the 0.7–0.8 range, suggesting that positional information reduces ordering uncertainty in tasks with more complex logic and longer execution trajectories.

$\bm{\lambda}$ shows only minor differences across the two subsets. On the Alg subset, performance increases with larger values and peaks at 0.9, indicating that strengthening temporal consistency constraints helps the model to capture long-range dependencies. On the HC subset, accuracy is high when $\lambda$ is small and decreases as the weight increases, suggesting that temporal consistency mainly acts as a regularization factor rather than directly driving the attribution.

\end{document}